%% file: main.tex
\documentclass[10pt,twocolumn,letterpaper]{article}

\usepackage[applications]{wacv}             % CAMERA-READY version


\usepackage{booktabs}
\usepackage{multirow}
\usepackage{makecell}
\usepackage{listings}
\usepackage{placeins}
\newcommand{\cond}[1]{\textsf{\textbf{#1}}}

\definecolor{wacvblue}{rgb}{0.21,0.49,0.74}
\usepackage[breaklinks,colorlinks,allcolors=wacvblue]{hyperref}

\def\wacvPaperID{24}
\def\confName{CV4EO@WACV}
\def\confYear{2027}

\title{Beyond Before-and-After Images: Evaluating Temporal Grounding\\in a Lightweight VLM for Disaster Assessment}
\author{Shubh Garg \qquad Pratham Arora \qquad Debabrata Ghosh\\
Thapar Institute of Engineering \& Technology\\
{\tt\small shubhgarg265@gmail.com, parora\char`\_be24@thapar.edu, debabrata.ghosh@thapar.edu}
}

\begin{document}

\maketitle

\input{sec/0_abstract}
\input{sec/1_intro}
\input{sec/2_related}
\input{sec/3_protocol}
\input{sec/4_results}
\input{sec/5_discussion}
\input{sec/6_conclusion}

{
    \small
    \bibliographystyle{ieeenat_fullname}
    \bibliography{main}
}

\input{sec/7_supp}

\end{document}

%% file: sec/0_abstract.tex
\begin{abstract}
Bi-temporal satellite imagery is central to disaster assessment: deciding what was damaged requires comparing a pre-event scene with its post-event counterpart. A vision-language model (VLM) can nonetheless appear to answer change questions while relying mainly on the post-event image or on the prompt's temporal framing. We present a controlled diagnostic of temporal grounding and apply it to one lightweight VLM, InternVL3-2B, on DisasterM3: 1,659 optical and SAR image pairs from 41 events and nine disaster types, in event-disjoint partitions, with 9,031 reasoning questions. We compare post-only input, the true pre/post pair, the post image repeated twice, reversed image order, and an unrelated pre-event image. The true pair scores $+0.5$ accuracy points above the repeated-post control (95\% paired bootstrap CI $[-0.2, +1.2]$), and reversing the order changes accuracy by $-0.2$ points ($[-0.8, +0.5]$). Under this protocol we therefore find no measurable benefit from the content of the true pre-event image and little sensitivity to temporal order. The model is not blind to the first image, however: the true pair scores $2.5$ points above a pair whose pre-event image comes from another event ($[+1.6, +3.4]$). State-transition and multi-hop questions remain hard, and optical and SAR inputs show different profiles. We outline temporal scene graphs as a testable next step, not a validated solution, and argue that temporal benchmarks for Earth observation should report whether answers depend on the intended evidence, not only how often they are right.
\end{abstract}

%% file: sec/1_intro.tex
\section{Introduction}
\label{sec:intro}

Rapid damage assessment from Earth observation (EO) needs more than a description of the scene after an event: a system must compare observations across time, decide which structures changed, and tell damaged objects from unchanged ones. Building-damage benchmarks have made this comparison explicit, from optical xBD~\cite{gupta2019xbd} to the optical--SAR BRIGHT~\cite{chen2025bright} and the instruction-based DisasterM3~\cite{wang2025disasterm3}, and dedicated change-detection networks~\cite{daudt2018fully,chen2020levir,chen2022bit,bandara2022changeformer} build the comparison into their architecture. Remote-sensing VLMs~\cite{lobry2020rsvqa,kuckreja2024geochat,liu2024remoteclip,irvin2025teochat} instead receive both images in a prompt and are expected to compare them on their own~\cite{liu2024temporalvlm}. Instruction-tuned VLMs~\cite{liu2023llava} and their compact successors~\cite{zhu2025internvl3,wang2024qwen2vl} are attractive for resource-limited response settings, but strong single-image question answering does not imply that a model uses the second image.

\begin{figure*}[t]
  \centering
  \includegraphics[width=\textwidth]{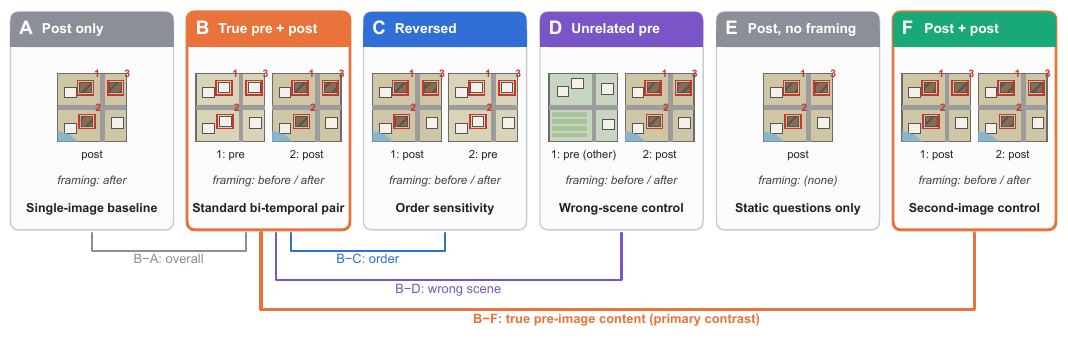}
  \caption{\textbf{Six input conditions and the contrasts between them.} Each card shows the images given to the model (schematic tiles, not dataset imagery), the temporal framing in the prompt, and the role of the condition. Numbered boxes sit at the same position on both images of a pair. The primary test of pre-image content is \cond{B}$-$\cond{F}: both conditions supply two images and the same before/after framing, and differ only in whether the first image is the true pre-event scene. Full definitions in \cref{tab:conditions}.}
  \label{fig:conditions}
\end{figure*}

Accuracy alone does not reveal which evidence a model used. A model may answer from the post-event scene alone, from question priors~\cite{goyal2017vqav2,agrawal2018dont}, from a bias toward ``yes''~\cite{li2023pope}, or from the mere presence of two images~\cite{fu2024blink,wang2025muirbench}; all are forms of shortcut behavior~\cite{geirhos2020shortcut}. We therefore measure temporal grounding by varying the input while holding the question fixed (\cref{fig:conditions}): post only (\cond{A}), true pair (\cond{B}), reversed pair (\cond{C}), unrelated pre-event image (\cond{D}), post only without temporal framing (\cond{E}, static questions), and the post image shown twice under the same before/after framing (\cond{F}). The usual comparison, \cond{B}$-$\cond{A}, mixes two effects: the content of the pre-event image, and the mere addition of a second image with before/after wording. \cond{B}$-$\cond{F} holds the second effect fixed and isolates the first.

Our goal is diagnostic rather than architectural. We contribute (1) a multi-condition protocol for auditing whether a VLM's answers depend on pre-event evidence, with \cond{B}$-$\cond{F} as the declared primary contrast; (2) its application to InternVL3-2B~\cite{zhu2025internvl3} on an event-disjoint set of 1,659 optical and SAR pairs and 9,031 reasoning questions, with clustered uncertainty intervals; (3) category-, sensor-, and template-level analyses, including an answer-bias probe; and (4) a falsifiable plan for temporal scene-graph grounding, presented as future work. Unlike temporal EO evaluations that report accuracy on change questions~\cite{yuan2022cdvqa,irvin2025teochat}, we hold each question and its post-event image fixed and vary only the accompanying evidence, so differences between conditions can be attributed to the input rather than to the question set. The protocol is model-agnostic, but the findings reported here are for one model.

The supplementary material gives an FAQ (\cref{app:faq}), full definitions, extended results, the proposed graph method, and the run record.

%% file: sec/2_related.tex
\section{Related Work}
\label{sec:related}

\noindent\textbf{Disaster EO and change detection.}
Supervised change detectors pair a Siamese encoder~\cite{daudt2018fully} with attention~\cite{chen2020levir} or transformers~\cite{chen2022bit,bandara2022changeformer}, so the comparison of the two dates is part of the model. Damage datasets supply the labels: xBD~\cite{gupta2019xbd} for optical imagery and BRIGHT~\cite{chen2025bright} for optical pre-event and SAR post-event pairs. DisasterM3~\cite{wang2025disasterm3} adds instruction-style questions over such pairs, which is what lets a general VLM be evaluated at all. Self-supervised EO backbones also model time explicitly~\cite{cong2022satmae}.

\noindent\textbf{Temporal VLMs for EO.}
Remote-sensing VQA~\cite{lobry2020rsvqa} and grounded RS assistants~\cite{kuckreja2024geochat,liu2024remoteclip} are mostly single-date. Change-aware question answering~\cite{yuan2022cdvqa}, change-interpretation agents~\cite{liu2024changeagent}, and temporal EO assistants~\cite{irvin2025teochat} take image sequences, and a recent survey~\cite{liu2024temporalvlm} catalogues this line. These works report accuracy on temporal questions; they rarely test whether the earlier image changed the answer.

\noindent\textbf{Evidence use and shortcuts.}
VQA models are known to answer from language priors~\cite{goyal2017vqav2,agrawal2018dont}, hallucinate objects~\cite{li2023pope}, and lose accuracy on multi-image perception~\cite{fu2024blink,wang2025muirbench}; prompt formatting alone can shift results~\cite{sclar2024quantifying}. We bring these concerns to bi-temporal EO, using input controls rather than new data to separate evidence use from framing.

%% file: sec/3_protocol.tex
\section{Data and Evaluation Protocol}
\label{sec:protocol}

\noindent\textbf{Data.}
We use the DisasterM3 pairs \cite{wang2025disasterm3} that were available locally with damage masks and readable images: 1,659 pairs from 41 events and nine disaster types, 1,235 optical and 424 SAR. Events are split into disjoint train, development, and test partitions (254 / 260 / 1,145 pairs). The model is evaluated zero-shot and nothing is fit to any partition, so headline numbers pool all three; we label them as a diagnostic and have not yet repeated the contrasts on the test partition alone. \cref{tab:setup} summarizes the setup and \cref{tab:filtering} the exclusions behind the 1,659 pairs.

\noindent\textbf{Questions.}
From masks and labels we generate 10,693 questions: 1,662 static and 9,031 reasoning questions in five categories (temporal, state transition, relational, multi-hop, causal; \cref{tab:main}). Within each split, yes/no answers are balanced 50/50 and no other answer class exceeds 1.5$\times$ the rarest. Images are resized to 448\,px, and numbered boxes are drawn at the same positions on both images of a pair. The generated questions have not been manually audited, and neither the resolution nor the boxes were ablated (\cref{sec:conclusion}). Each question has three neutral wordings; wording 1 is used for every condition and wordings 2--3 are also run under \cond{B} (\cref{tab:wording}). ``Causal'' questions are dataset-defined and observational; they are not intervention-based causal ground truth~\cite{pearl2009causality}.

\noindent\textbf{Conditions.}
\cref{fig:conditions} and \cref{tab:conditions} define the conditions. \cond{B}$-$\cond{F} is the primary contrast (\cref{sec:intro}); \cond{B}$-$\cond{C} probes order, \cond{B}$-$\cond{D} compares the true pre-event image with one from another event, \cond{F}$-$\cond{A} measures adding a second image with paired framing, and \cond{E} (no framing) is a static-only control.

\noindent\textbf{Model.}
OpenGVLab/InternVL3-2B-hf~\cite{zhu2025internvl3} runs in bfloat16 with greedy decoding and at most 10 new tokens, through PyTorch~\cite{paszke2019pytorch} and Transformers~\cite{wolf2020transformers} on a 6\,GB laptop GPU. All 72,846 calls returned a parseable answer (\cref{tab:runrecord}).

\noindent\textbf{Statistics.}
Questions from one pair are not independent, so confidence intervals use 2,000 paired bootstrap resamples~\cite{efron1993bootstrap} of image pairs within events~\cite{field2007bootstrapping}. We report paired deltas in accuracy points (pp) with 95\% intervals. An interval that includes zero is read as ``no difference established'', never as equivalence, which would need a pre-declared margin~\cite{lakens2017equivalence}. Per-category tests are exploratory (\cref{tab:stats}).

%% file: sec/4_results.tex
\section{Results}
\label{sec:results}

\begin{table}[t]
\centering
\footnotesize
\caption{\textbf{Core results.} Top: paired contrasts over all 9,031 reasoning questions, in accuracy points with 95\% clustered bootstrap CIs. Bottom: accuracy under \cond{B} by category; chance is the mean of 1/(number of options).}
\label{tab:main}
\setlength{\tabcolsep}{4pt}
\begin{tabular}{@{}llrl@{}}
\toprule
Contrast & Isolates & $\Delta$ (pp) & 95\% CI \\
\midrule
\cond{B}$-$\cond{F} & pre-image content (primary) & $+0.5$ & $[-0.2, +1.2]$ \\
\cond{B}$-$\cond{A} & whole input package & $0.0$ & $[-0.9, +0.9]$ \\
\cond{B}$-$\cond{C} & temporal order & $-0.2$ & $[-0.8, +0.5]$ \\
\cond{F}$-$\cond{A} & second image + framing & $-0.5$ & $[-1.4, +0.4]$ \\
\cond{B}$-$\cond{D} & true vs.\ unrelated pre & $+2.5$ & $[+1.6, +3.4]$ \\
\midrule
Category & $n$ & Acc.\ \cond{B} & Chance \\
\midrule
Static & 1,662 & 0.534 & 0.39 \\
Temporal & 2,106 & 0.456 & 0.42 \\
State transition & 1,319 & 0.389 & 0.27 \\
Relational & 3,293 & 0.582 & 0.37 \\
Multi-hop & 1,221 & 0.326 & 0.29 \\
Causal & 1,092 & 0.533 & 0.25 \\
\bottomrule
\end{tabular}
\end{table}

\begin{figure}[t]
\centering
\includegraphics[width=\linewidth]{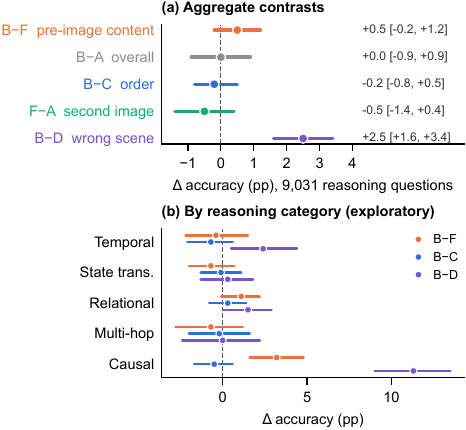}
\caption{\textbf{Paired contrasts with 95\% CIs.} (a) Aggregate over reasoning questions. (b) Per category; 30 such tests are run (\cref{tab:deltas}), so isolated intervals excluding zero are exploratory. The causal effect of \cond{B}$-$\cond{D} reflects scene plausibility questions (\cref{tab:templates}).}
\label{fig:contrasts}
\end{figure}

\noindent\textbf{The true pre-event image adds no measurable accuracy.}
Against the repeated-post control, the true pair scores $+0.5$\,pp, with an interval that contains zero (\cref{tab:main}, \cref{fig:contrasts}a). Reversing the order changes accuracy by $-0.2$\,pp, and the conventional \cond{B}$-$\cond{A} contrast is exactly zero. Under this protocol, the content and order of the first image do not measurably change aggregate accuracy. The model is not blind to the first image, however: replacing the true pre-event image with one from another event costs $2.5$\,pp, an interval that excludes zero.

\noindent\textbf{Where the second image matters.}
Category effects are small except in causal questions, where both \cond{B}$-$\cond{F} ($+3.2$\,pp) and \cond{B}$-$\cond{D} ($+11.3$\,pp) exclude zero (\cref{fig:contrasts}b; all categories and conditions in \cref{tab:percat,tab:deltas}). Template-level results locate the effect: \texttt{plausible\_event} drops from 0.461 to 0.326 with an unrelated pre-image (\cref{tab:templates}), which fits a model that infers the event type from overall scene context rather than comparing objects across time. Multi-hop accuracy stays within 5.5\,pp of chance and state-transition accuracy below 0.40 under every condition, and reasoning depth matters more than the input condition: three-step questions score 0.326 against 0.501--0.521 for one or two steps (\cref{tab:wording}).

\begin{figure}[t]
\centering
\includegraphics[width=\linewidth]{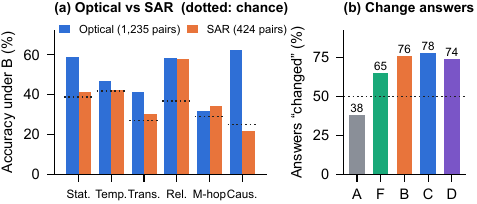}
\caption{\textbf{Sensor and answer bias.} (a) Accuracy under \cond{B} for optical and SAR pairs (Stat.\ static, Temp.\ temporal, Trans.\ state transition, Rel.\ relational, M-hop multi-hop, Caus.\ causal); dotted lines mark chance. (b) Share of ``changed'' answers to \texttt{changed\_yes\_no} by condition, where the true rate is 50\%.}
\label{fig:sensorbias}
\end{figure}

\noindent\textbf{Optical and SAR differ.}
Under \cond{B}, optical pairs score higher on static (0.588 vs.\ 0.415), state-transition (0.413 vs.\ 0.304), and especially causal questions (0.622 vs.\ 0.220, below chance for SAR), while relational and multi-hop scores are similar (\cref{fig:sensorbias}a). Sensor is confounded with event, disaster type, and question mix (\cref{tab:sensor}), so we report these differences descriptively.

\noindent\textbf{The second image mainly shifts the answer prior.}
On the balanced \texttt{changed\_yes\_no} template, the model answers ``changed'' for 38\% of post-only inputs but 65--78\% whenever two images are shown, whether or not the first is the true pre-event scene (\cref{fig:sensorbias}b). Under \cond{B}, the rate of ``changed'' answers is only $+4.2$\,pp higher for truly changed than for unchanged cases ($[-0.1, +8.6]$; \cref{tab:yesno}). A change-implying wording removes ``no change'' answers entirely and lowers building-transition accuracy by 5.7\,pp (\cref{tab:yesno}). Answer rates thus follow input format and wording more closely than the actual change, echoing the prompt sensitivity reported for language models~\cite{sclar2024quantifying}.

%% file: sec/5_discussion.tex
\section{Discussion and Next Step}
\label{sec:discussion}

\noindent\textbf{The pre-event image goes largely unused.}
In our setting the true pre-event image was no more useful than a copy of the post-event image, order barely mattered, and a second image mainly shifted answers toward ``changed''. One could read this as evidence that the pre-event image is simply unnecessary for these questions. Three observations argue against that reading. First, the change questions are not being solved from the post-event image either: on the balanced \texttt{changed\_yes\_no} template, post-only input gives no measurable discrimination between changed and unchanged cases (truth-conditioned gap $+0.016$, $[-0.035,+0.065]$), and the true pair barely improves on it ($+0.042$, $[-0.001,+0.086]$; \cref{tab:yesno}). If the post image sufficed, \cond{A} would already separate the two classes. Second, on templates whose labels are defined relative to the pre-event state, accuracy sits near or below the majority baseline under every condition: \texttt{building\_transition} scores 0.349 under \cond{B} against a majority baseline of 0.375, and \texttt{which\_changed} scores 0.330 under \cond{B} but 0.355 with the order reversed (\cref{tab:templates}). A model that compared the two dates should do better with the correct order, not worse. Third, transition answers follow the wording rather than the images: a change-implying prompt eliminates every ``no change'' answer (22\%$\rightarrow$0\%) and lowers building-transition accuracy by 5.7\,pp (\cref{tab:yesno}). Together, these results indicate that the information needed to answer is not being extracted from either image, and that the pre-event image, the evidence designed to resolve these questions, goes unused. The model is not insensitive to the first image: an unrelated scene costs 2.5\,pp, concentrated in event-plausibility questions (\cref{tab:templates}), a pattern that fits coarse scene-context matching rather than object-level comparison across time.

\noindent\textbf{Model or dataset?}
This argument rests on a few templates, and the protocol alone cannot say whether the shortcut belongs to the model or to the questions. We have no upper bound (human, oracle, or fine-tuned model) showing that the task is solvable with the pre-event image, and no pre-only condition. A null \cond{B}$-$\cond{C} is also ambiguous: the model may ignore order, or the direction of time may be inferable from physical cues such as debris or floodwater regardless of order. A constructed control set would separate these readings: synthetic pairs in which objects are added, removed, or resized between frames, with questions such as ``Are there more circles after the event than before?'' that cannot be answered without both images in the correct order. Failure on such a set would locate the limitation in the model; success would point to the DisasterM3 questions.

\noindent\textbf{Making the comparison explicit.}
A temporal scene graph~\cite{johnson2015scenegraph,ji2020actiongenome} would hold entities, their states at $t_0$ and $t_1$, spatial relations, and transition edges such as \textit{intact}$\rightarrow$\textit{damaged}, each with confidence and provenance (\cref{fig:tsg}, \cref{lst:tsg}). We intend the same VLM to build the graph in a two-stage prompt, first listing the marked entities and their states in each image, then answering from its own graph and the images. No external model adds capability, so the comparison with \cond{B} stays fair, and each answer becomes auditable: a correct answer that cites no transition edge is suspect for the same reason a correct answer under \cond{F} is. Graphs from annotations (oracle) or a separate detector are diagnostics only: they test whether the reasoning stage can use temporal facts, but move the hard part, building the graph from the pair, outside the model. Corrupted- and reversed-graph controls mirror \cond{D} and \cond{C} (\cref{tab:ablation}). The prediction is falsifiable: if explicit comparison is what is missing, self-built graphs should raise transition and change-question accuracy and restore order sensitivity, while corrupted and reversed graphs should not. No graph experiment has been run for this paper.

%% file: sec/6_conclusion.tex
\section{Limitations and Conclusion}
\label{sec:conclusion}

All results come from one 2B-parameter VLM, one question generator, and one prompt protocol; they do not show that lightweight VLMs in general neglect temporal evidence. Repeating the protocol on a second small model (\eg, Qwen2-VL~\cite{wang2024qwen2vl}) and a larger one is the first extension. The task lacks an upper bound and a pre-only condition, and model- and data-side explanations are not yet separated (\cref{sec:discussion}). The generated questions were not manually audited, and the 448\,px resizing and numbered boxes, which may remove fine damage detail or distract the model, were not ablated (\cref{app:faq}). Optical and SAR subsets differ in events and question mix, headline numbers pool three event-disjoint partitions, causal labels are observational, and a non-significant difference is not proof of equivalence. Recorded and missing provenance is listed in \cref{tab:runrecord}, following datasheet practice~\cite{gebru2021datasheets}.

We presented a controlled test of whether a lightweight VLM uses the pre-event image in bi-temporal disaster imagery. With the repeated-post control, the true pre/post pair showed no measurable aggregate gain, reversing the order changed little, and a second image mostly shifted the answer prior. We recommend that temporal EO benchmarks report a repeated-post control alongside accuracy, and we propose self-built temporal scene graphs as the next representation to test under the same event-disjoint protocol.

%% file: sec/7_supp.tex
% Supplementary material, appended to main.tex after the references.
\clearpage
\setcounter{section}{0}\setcounter{table}{0}\setcounter{figure}{0}\setcounter{lstlisting}{0}
\renewcommand{\thesection}{S\arabic{section}}
\renewcommand{\thetable}{S\arabic{table}}
\renewcommand{\thefigure}{S\arabic{figure}}
\renewcommand{\thelstlisting}{S\arabic{lstlisting}}
\maketitlesupplementary

\noindent References of the form Sec.~1--6, Fig.~1--3 and Tab.~1 point to the main paper; S-numbered items are in this supplement.

% ---------------------------------------------------------------------------
\FloatBarrier
\section{Frequently Asked Questions}
\label{app:faq}

\noindent\textbf{Q1. What is the central contribution?}
A controlled diagnostic of whether a lightweight VLM uses paired pre- and post-disaster satellite imagery. The value lies in the controls (repeating the post image, reversing order, and pairing an unrelated pre-event image), not in a new architecture (\cref{sec:protocol}).

\noindent\textbf{Q2. Does the study show the model cannot understand disaster change?}
No. Under the tested prompts, data, model, and protocol, the correct pre-event image did not produce a reliable aggregate improvement over the repeated-post control (\cref{tab:main}). This does not show that every VLM fails, or that every form of temporal grounding is ineffective.

\noindent\textbf{Q3. Why compare the true pair with a repeated post image?}
\cond{F} matches \cond{B} in image count and temporal framing, so \cond{B}$-$\cond{F} separates the value of genuine pre-event content from the effect of a second image or of before/after wording. \cond{B}$-$\cond{A} mixes these effects.

\noindent\textbf{Q4. Why include reversed-order and unrelated-pre controls?}
Reversal (\cond{C}) tests sensitivity to temporal order. An unrelated pre-event image (\cond{D}) tests whether the model benefits from any extra image or specifically from relevant pre-event evidence. Together they make the grounding claim more diagnostic than a post-only versus pair comparison.

\noindent\textbf{Q5. Why does the true pair beat an unrelated pair?}
\cond{B}$-$\cond{D} is positive ($+2.5$\,pp, $[+1.6,+3.4]$), but it is concentrated in causal questions and the \texttt{plausible\_event} template (\cref{tab:deltas,tab:templates}). Mismatched scene context, rather than object-level comparison, is a sufficient explanation; it does not by itself show robust temporal reasoning.

\noindent\textbf{Q6. Are the results evidence of causal reasoning?}
Not in the intervention-based sense~\cite{pearl2009causality}. The causal category scores answers to dataset questions; observational pre/post imagery cannot establish physical causation. We separate observed state changes, plausible explanations, and verified causal ground truth (\cref{fig:tsg}).

\noindent\textbf{Q7. Has the temporal scene-graph method been validated?}
No. It is a motivated next step (\cref{sec:discussion}, \cref{app:tsg}). No graph-grounded experiment has been run, and we make no claim that it improves accuracy.

\noindent\textbf{Q8. Who builds the scene graph, the VLM or another model?}
The proposed method has the same VLM build it: a two-stage prompt first asks the model to list the marked entities and their states in each image, then to answer from that graph and the images. No external model adds capability, so the comparison with \cond{B} stays fair. Graphs built from annotations (oracle) or by a separate detector are diagnostics only: they show whether the reasoning stage can use temporal facts, but they move the hard part, building the graph from the pair, outside the model (\cref{tab:ablation}).

\noindent\textbf{Q9. Do the findings hold for other VLMs?}
Unknown. Every result comes from InternVL3-2B, and the protocol has not yet been repeated on a second model. The conditions and contrasts are model-agnostic, so the first extension is a second small VLM (\eg, Qwen2-VL~\cite{wang2024qwen2vl}) and one larger model, which would show whether the effect depends on model size. Fine-tuning is a separate question: it asks whether training raises scores, not whether available temporal evidence is used, and it can introduce data and forgetting effects that obscure this diagnostic.

\noindent\textbf{Q10. Is the shortcut a property of the model or of the dataset?}
The protocol alone cannot tell. The questions may be answerable from the post-event image, or too hard for any model; and a null order effect could mean that the model ignores order or that the direction of time is visible from physical cues. Three additions would separate these readings: a pre-only condition, an upper bound from a human or oracle reader, and a constructed control set of synthetic pairs (objects added, removed, or resized) whose questions cannot be answered without both images in the correct order (\cref{sec:discussion}). The argument in the main paper that the post image does not suffice rests mainly on \texttt{changed\_yes\_no} and the transition templates.

\noindent\textbf{Q11. Could question quality, resizing, or the drawn boxes explain the results?}
Possibly, in part. The questions are generated automatically from masks and labels and have not been manually audited, so some may be ambiguous. Resizing to 448\,px may remove fine damage detail, and the numbered boxes may distract the model; neither was ablated. They apply equally to every condition, so they cannot manufacture a difference between conditions, but they could hide one: if fine damage detail is lost at 448\,px, the pre-event image has less to contribute, which would shrink \cond{B}$-$\cond{F}. A manual audit of a question sample and a resolution and no-box ablation are planned.

\noindent\textbf{Q12. Which split do the headline numbers use?}
All three event-disjoint partitions, pooled (\cref{tab:setup}). The model is zero-shot and nothing is fit to any partition, so we label the numbers as a diagnostic. A test-only (1,145-pair) repetition has not yet been run.

\noindent\textbf{Q13. How were exclusions handled?}
\cref{tab:filtering} lists exclusion counts as recorded by the data loader. The counts mix image- and pair-level units and may overlap, so we do not sum them or present them as a mutually exclusive flow.

\noindent\textbf{Q14. Are the failure-mechanism counts ground-truth error categories?}
No. They are assigned by rules, and the planned 50-case manual audit has not been completed. We report them only as heuristic labels (\cref{tab:failures}) and base no claim on them.

\noindent\textbf{Q15. How should optical and SAR results be read?}
Separately, because their visual characteristics and performance differ (\cref{fig:sensorbias}a). Sensor is confounded with event composition and question mix, and several disaster types have few pairs (\cref{tab:sensor}), so we draw no causal conclusion about sensor type.

\noindent\textbf{Q16. What is the strongest defensible takeaway?}
For the tested model and setup, supplying a pre/post pair is not enough to reliably improve temporal reasoning. This motivates testing explicit, question-relevant temporal structure; it does not yet show that a graph will solve the problem.

% ---------------------------------------------------------------------------
\FloatBarrier
\section{Setup, Input Conditions, and Prompts}
\label{app:conditions}

\begin{table}[!htbp]
\centering
\footnotesize
\caption{\textbf{Evaluation setup.} Data, model, and inference settings for all conditions; the complete run record is given in \cref{tab:runrecord}.}
\label{tab:setup}
\setlength{\tabcolsep}{4pt}
\begin{tabular}{@{}ll@{}}
\toprule
Pairs (optical / SAR) & 1,659 (1,235 / 424) \\
Events / disaster types & 41 / 9 \\
Event-disjoint split (pairs) & 254 / 260 / 1,145 (pooled) \\
Questions (static / reasoning) & 10,693 (1,662 / 9,031) \\
Model & InternVL3-2B, bf16, greedy, $\le$10 tokens \\
Input & 448\,px; numbered boxes on both images \\
Calls / errors / unparseable & 72,846 / 0 / 0 \\
Software & PyTorch 2.11.0, Transformers 5.12.1 \\
Intervals & 2,000 resamples of pairs within events \\
\bottomrule
\end{tabular}
\end{table}

\begin{table*}[!tbp]
\centering
\footnotesize
\caption{\textbf{Condition registry.} All conditions use question wording 1. \cond{E} applies to static questions only; \cond{C}, \cond{D} and \cond{F} to reasoning questions only.}
\label{tab:conditions}
\begin{tabular}{@{}lp{0.2\textwidth}p{0.33\textwidth}p{0.12\textwidth}p{0.19\textwidth}@{}}
\toprule
ID & Image input & Prompt framing & Questions & Role \\
\midrule
\cond{A} & post-event image once & ``This image was taken after the disaster'' & static + reasoning & single-image baseline \\
\cond{B} & pre-event, then post-event & ``The first image was taken before the disaster and the second after it'' & static + reasoning & primary bi-temporal condition \\
\cond{C} & post-event, then pre-event & same statement as \cond{B} & reasoning & order sensitivity \\
\cond{D} & pre-event image from another event, then correct post & same statement as \cond{B} & reasoning & wrong-scene control \\
\cond{E} & post-event image once & no temporal framing & static & static framing control \\
\cond{F} & post-event image twice & same statement as \cond{B} & reasoning & second-image control \\
\bottomrule
\end{tabular}
\end{table*}

\cref{tab:conditions} gives the exact framing. Questions have three neutral wordings plus, for transition questions, one change-implying wording that is scored separately (\cref{tab:yesno}). Balancing is applied within each split. Primary contrast: \cond{B}$-$\cond{F}. Secondary: \cond{B}$-$\cond{A}, \cond{B}$-$\cond{C}, \cond{B}$-$\cond{D}, \cond{F}$-$\cond{A}. These contrasts answer different questions and are not interchangeable.

% ---------------------------------------------------------------------------
\FloatBarrier
\section{Statistical Protocol}
\label{app:stats}

\begin{table}[!htbp]
\centering
\footnotesize
\caption{\textbf{Statistical protocol and reporting rules.}}
\label{tab:stats}
\setlength{\tabcolsep}{3pt}
\begin{tabular}{@{}p{0.33\linewidth}p{0.62\linewidth}@{}}
\toprule
Confidence intervals & 95\%, 2,000 bootstrap resamples~\cite{efron1993bootstrap} \\
Resampling unit & image pair, resampled within event~\cite{field2007bootstrapping} \\
Paired comparison & deltas computed on the same questions \\
Averaging & micro-average over questions \\
Multiplicity & 30 per-category tests; treated as exploratory \\
Chance & mean of 1/(number of options); template-specific \\
Majority baseline & most frequent exact answer string \\
Equivalence & not claimed; no pre-declared margin~\cite{lakens2017equivalence} \\
\bottomrule
\end{tabular}
\end{table}

\cref{tab:stats} summarises the protocol. The interval type (percentile or BCa) was not recorded in the run report and will be stated with the released code (\cref{tab:runrecord}).

% ---------------------------------------------------------------------------
\FloatBarrier
\section{Extended Results}
\label{app:results}

\begin{table*}[!tbp]
\centering
\footnotesize
\caption{\textbf{Accuracy by category and condition} (95\% CIs). ``--'' marks conditions not run for that category.}
\label{tab:percat}
\setlength{\tabcolsep}{4pt}
\begin{tabular}{@{}lrrcccccc@{}}
\toprule
Category & $n$ & Chance & \cond{A} & \cond{F} & \cond{B} & \cond{C} & \cond{D} & \cond{E} \\
\midrule
Static & 1,662 & 0.39 & 0.521 [.500,.541] & -- & 0.534 [.513,.555] & -- & -- & 0.530 [.509,.551] \\
Temporal & 2,106 & 0.42 & 0.445 [.425,.467] & 0.460 [.439,.480] & 0.456 [.434,.476] & 0.463 [.443,.483] & 0.432 [.411,.453] & -- \\
State transition & 1,319 & 0.27 & 0.372 [.347,.397] & 0.396 [.370,.421] & 0.389 [.361,.413] & 0.390 [.363,.414] & 0.386 [.360,.412] & -- \\
Relational & 3,293 & 0.37 & 0.595 [.579,.610] & 0.571 [.555,.586] & 0.582 [.566,.598] & 0.579 [.563,.594] & 0.567 [.551,.583] & -- \\
Multi-hop & 1,221 & 0.29 & 0.343 [.318,.370] & 0.333 [.309,.358] & 0.326 [.300,.353] & 0.328 [.302,.352] & 0.326 [.302,.353] & -- \\
Causal & 1,092 & 0.25 & 0.514 [.489,.538] & 0.501 [.477,.525] & 0.533 [.510,.556] & 0.538 [.513,.562] & 0.420 [.396,.444] & -- \\
\bottomrule
\end{tabular}
\end{table*}

\begin{table*}[!tbp]
\centering
\footnotesize
\caption{\textbf{Per-category paired contrasts} (accuracy, 95\% CIs). With 30 category-level tests, one or two intervals excluding zero are expected by chance.}
\label{tab:deltas}
\setlength{\tabcolsep}{4pt}
\begin{tabular}{@{}lccccc@{}}
\toprule
Category & \cond{B}$-$\cond{F} (pre content) & \cond{B}$-$\cond{C} (order) & \cond{F}$-$\cond{A} (2nd image) & \cond{B}$-$\cond{A} (overall) & \cond{B}$-$\cond{D} (wrong scene) \\
\midrule
Temporal & $-$.004 [$-$.022,+.015] & $-$.007 [$-$.021,+.006] & +.015 [$-$.009,+.038] & +.010 [$-$.014,+.035] & +.024 [+.005,+.044] \\
State transition & $-$.007 [$-$.020,+.007] & $-$.001 [$-$.013,+.011] & +.024 [+.002,+.045] & +.017 [$-$.006,+.039] & +.003 [$-$.013,+.018] \\
Relational & +.011 [$-$.001,+.022] & +.003 [$-$.008,+.014] & $-$.024 [$-$.038,$-$.008] & $-$.013 [$-$.029,+.004] & +.015 [.000,+.029] \\
Multi-hop & $-$.007 [$-$.028,+.012] & $-$.002 [$-$.020,+.016] & $-$.010 [$-$.028,+.009] & $-$.017 [$-$.042,+.006] & .000 [$-$.024,+.022] \\
Causal & +.032 [+.016,+.048] & $-$.005 [$-$.017,+.006] & $-$.013 [$-$.031,+.005] & +.019 [$-$.002,+.041] & +.113 [+.090,+.135] \\
\midrule
All reasoning & +.005 [$-$.002,+.012] & $-$.002 [$-$.008,+.005] & $-$.005 [$-$.014,+.004] & .000 [$-$.009,+.009] & +.025 [+.016,+.034] \\
\bottomrule
\end{tabular}
\end{table*}

\begin{table*}[!tbp]
\centering
\footnotesize
\caption{\textbf{Template-level accuracy.} Majority: always the most frequent exact answer string. For direction templates the answer names a reference box, so 1-in-4 is a fairer chance level. Top share: frequency of the model's most common answer under \cond{B}.}
\label{tab:templates}
\setlength{\tabcolsep}{4.5pt}
\begin{tabular}{@{}lrcccccccc@{}}
\toprule
Template & $n$ & Majority & \cond{A} & \cond{F} & \cond{B} & \cond{C} & \cond{D} & Bal.\ acc.\ (\cond{B}) & Top share (\cond{B}) \\
\midrule
\texttt{object\_type} & 752 & 0.500 & 0.507 & -- & 0.477 & -- & -- & 0.507 & 72\% \\
\texttt{road\_present} & 910 & 0.500 & 0.533 & -- & 0.580 & -- & -- & 0.549 & 80\% \\
\texttt{changed\_yes\_no} & 1,388 & 0.500 & 0.508 & 0.531 & 0.521 & 0.519 & 0.490 & 0.521 & 76\% \\
\texttt{which\_changed} & 718 & 0.241 & 0.325 & 0.323 & 0.330 & 0.355 & 0.319 & 0.321 & 36\% \\
\texttt{building\_transition} & 995 & 0.375 & 0.315 & 0.347 & 0.349 & 0.349 & 0.356 & -- & 48\% \\
\texttt{road\_transition} & 324 & 0.438 & 0.549 & 0.546 & 0.512 & 0.515 & 0.478 & -- & 41\% \\
\texttt{adjacent\_yes\_no} & 760 & 0.500 & 0.549 & 0.545 & 0.551 & 0.533 & 0.561 & 0.551 & 65\% \\
\texttt{nearest\_building} & 1,242 & 0.254 & 0.738 & 0.729 & 0.759 & 0.753 & 0.738 & 0.754 & 24\% \\
\texttt{relative\_position} & 1,291 & 0.070 & 0.484 & 0.435 & 0.428 & 0.438 & 0.407 & 0.420 & 14\% \\
\texttt{changed\_direction} & 755 & 0.078 & 0.265 & 0.257 & 0.244 & 0.246 & 0.240 & 0.234 & 12\% \\
\texttt{nearest\_to\_changed} & 466 & 0.227 & 0.470 & 0.457 & 0.459 & 0.459 & 0.466 & 0.469 & 27\% \\
\texttt{plausible\_event} & 902 & 0.224 & 0.436 & 0.419 & 0.461 & 0.469 & 0.326 & 0.482 & 30\% \\
\texttt{why\_road\_inaccessible} & 190 & 0.568 & 0.884 & 0.889 & 0.874 & 0.863 & 0.868 & 0.855 & 68\% \\
\bottomrule
\end{tabular}
\end{table*}

\begin{table}[t]
\centering
\footnotesize
\caption{\textbf{Yes/no answer bias.} Gap $=P(\text{Yes}\mid\text{true Yes})-P(\text{Yes}\mid\text{true No})$ with 95\% CI; a gap near zero means weak discrimination.}
\label{tab:yesno}
\setlength{\tabcolsep}{3pt}
\begin{tabular}{@{}llrrl@{}}
\toprule
Template & Cond. & $n$ & Yes & Gap [95\% CI] \\
\midrule
\multirow{5}{*}{\texttt{changed\_yes\_no}} & \cond{A} & 1,388 & 38\% & +.016 [$-$.035,+.065] \\
 & \cond{F} & 1,388 & 65\% & +.062 [+.014,+.112] \\
 & \cond{B} & 1,388 & 76\% & +.042 [$-$.001,+.086] \\
 & \cond{C} & 1,388 & 78\% & +.037 [$-$.005,+.083] \\
 & \cond{D} & 1,388 & 74\% & $-$.020 [$-$.066,+.024] \\
\midrule
\multirow{5}{*}{\texttt{adjacent\_yes\_no}} & \cond{A} & 760 & 72\% & +.097 [+.036,+.156] \\
 & \cond{F} & 760 & 29\% & +.089 [+.027,+.155] \\
 & \cond{B} & 760 & 35\% & +.103 [+.037,+.169] \\
 & \cond{C} & 760 & 42\% & +.066 [$-$.007,+.136] \\
 & \cond{D} & 760 & 42\% & +.121 [+.054,+.188] \\
\midrule
\multirow{3}{*}{\texttt{road\_present}} & \cond{E} & 910 & 84\% & +.099 [+.057,+.143] \\
 & \cond{A} & 910 & 88\% & +.066 [+.027,+.107] \\
 & \cond{B} & 910 & 80\% & +.160 [+.117,+.202] \\
\bottomrule
\end{tabular}
\end{table}

\begin{table}[t]
\centering
\footnotesize
\caption{\textbf{Wording robustness and reasoning depth} under \cond{B}. Top: accuracy over three neutral wordings (SD across wordings) and share of questions answered identically under all three. Bottom: accuracy by number of reasoning steps; the 3-step row is the multi-hop category.}
\label{tab:wording}
\setlength{\tabcolsep}{3pt}
\begin{tabular}{@{}lrcccc@{}}
\toprule
Category & $n$ & Mean (SD) & Worst & Best & Same \\
\midrule
Static & 1,662 & 0.545 (.010) & 0.534 & 0.558 & 80\% \\
Temporal & 2,106 & 0.456 (.003) & 0.453 & 0.460 & 67\% \\
State transition & 1,319 & 0.400 (.008) & 0.389 & 0.409 & 66\% \\
Relational & 3,293 & 0.572 (.008) & 0.563 & 0.582 & 63\% \\
Multi-hop & 1,221 & 0.323 (.023) & 0.293 & 0.351 & 48\% \\
Causal & 1,092 & 0.533 (.009) & 0.522 & 0.544 & 83\% \\
\midrule
Steps & $n$ & \multicolumn{2}{c}{Acc.\ \cond{B} [95\% CI]} & Chance & \\
\midrule
1 & 3,713 & \multicolumn{2}{c}{0.501 [.485,.515]} & 0.36 & \\
2 & 5,759 & \multicolumn{2}{c}{0.521 [.509,.533]} & 0.36 & \\
3 & 1,221 & \multicolumn{2}{c}{0.326 [.300,.353]} & 0.29 & \\
\bottomrule
\end{tabular}
\end{table}

\begin{table}[t]
\centering
\footnotesize
\caption{\textbf{Sensor and disaster-type breakdown} under \cond{B}. Explosion has eight pairs and is not a stable estimate. Per-event results (41 events) will be released with the code.}
\label{tab:sensor}
\setlength{\tabcolsep}{3.5pt}
\begin{tabular}{@{}lrccc@{}}
\toprule
Category & & Optical & SAR & \\
\midrule
Static & & 0.588 & 0.415 & \\
Temporal & & 0.469 & 0.424 & \\
State transition & & 0.413 & 0.304 & \\
Relational & & 0.582 & 0.580 & \\
Multi-hop & & 0.320 & 0.341 & \\
Causal & & 0.622 & 0.220 & \\
Pairs & & 1,235 & 424 & \\
\midrule
Disaster type & Pairs & Static & Reasoning & All \\
\midrule
Earthquake & 392 & 0.419 & 0.489 & 0.479 \\
Hurricane & 266 & 0.628 & 0.453 & 0.479 \\
Tornado & 255 & 0.484 & 0.483 & 0.483 \\
Wildfire & 240 & 0.579 & 0.516 & 0.528 \\
Volcanic eruption & 162 & 0.364 & 0.464 & 0.442 \\
Tsunami & 156 & 0.758 & 0.502 & 0.537 \\
Flood & 99 & 0.607 & 0.578 & 0.581 \\
Armed conflict & 81 & 0.538 & 0.373 & 0.398 \\
Explosion & 8 & 0.625 & 0.446 & 0.469 \\
\bottomrule
\end{tabular}
\end{table}

\cref{tab:percat,tab:deltas} give every category under every condition. \cref{tab:templates} breaks results down by question template, \cref{tab:yesno} reports answer bias, \cref{tab:wording} gives wording robustness and reasoning depth, and \cref{tab:sensor} splits by sensor and disaster type. \emph{Leading wording.} For transition questions under \cond{B}, replacing a neutral wording with a change-implying one lowers building-transition accuracy from 0.349 to 0.291 ($\Delta-0.057$ [$-0.093$, $-0.021$]) and removes ``no change'' answers (22\%$\rightarrow$0\%); road-transition accuracy moves from 0.512 to 0.494 ($\Delta-0.019$ [$-0.077$, $+0.043$]), with ``no change'' answers falling from 28\% to 0\%.

% ---------------------------------------------------------------------------
\FloatBarrier
\section{Data Filtering and Heuristic Failure Labels}
\label{app:filtering}

\begin{table}[!htbp]
\centering
\footnotesize
\caption{\textbf{Exclusions as recorded.} Units are mixed (images vs.\ pairs) and reasons may overlap, so the counts are not summed. No per-event cap was applied.}
\label{tab:filtering}
\setlength{\tabcolsep}{4pt}
\begin{tabular}{@{}lrl@{}}
\toprule
Reason & Count & Unit as recorded \\
\midrule
No damage mask & 861 & images \\
Image missing locally & 137 & images \\
No visible building & 208 & pairs \\
Fewer than 2 markable objects & 100 & pairs \\
Unreadable image file & 1 & image \\
\midrule
Usable pairs & 1,659 & pairs \\
\bottomrule
\end{tabular}
\end{table}

\begin{table}[!htbp]
\centering
\footnotesize
\caption{\textbf{Rule-assigned failure labels under \cond{B}}, \emph{not manually validated}. Shown for transparency only; no claim in the paper depends on them.}
\label{tab:failures}
\setlength{\tabcolsep}{4pt}
\begin{tabular}{@{}lr@{}}
\toprule
Rule-assigned label & Count \\
\midrule
Spatial relation failure & 1,378 \\
Ignored pre-image & 1,353 \\
Temporal order failure & 828 \\
Recognition failure & 775 \\
Hallucinated change & 516 \\
Single-hop / failed first hop & 195 \\
Other & 134 \\
Confused entity identity & 100 \\
Wrong state transition & 61 \\
Correct observation, wrong reasoning & 57 \\
Causal hallucination & 41 \\
\bottomrule
\end{tabular}
\end{table}

\cref{tab:filtering} lists the exclusions behind the 1,659 usable pairs, and \cref{tab:failures} the heuristic failure labels. A deduplicated filter flow in a single unit and a manual audit of the labels (sampling plan, annotators, agreement) are planned.

% ---------------------------------------------------------------------------
\FloatBarrier
\section{Proposed Temporal Scene-Graph Grounding}
\label{app:tsg}

\begin{figure*}[!tbp]
\centering
\includegraphics[width=\textwidth]{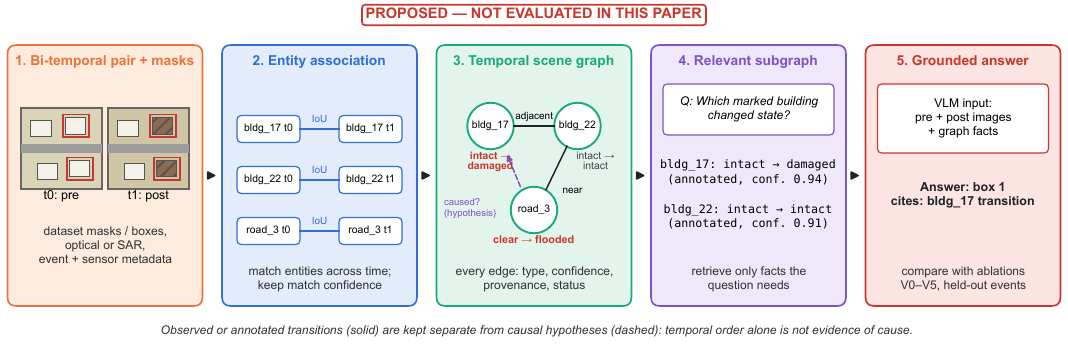}
\caption{\textbf{Proposed temporal scene-graph pipeline (not evaluated), shown for the annotation-derived (oracle) graph.} In the proposed self-built variant, the evaluated VLM performs steps 2--3 itself. (1) Start from the bi-temporal pair and dataset masks; (2) associate entities across time; (3) build a graph whose nodes carry states at $t_0$ and $t_1$ and whose edges record type, confidence, provenance, and status; (4) retrieve the subgraph a question needs; (5) answer from images plus graph facts. Causal links stay hypotheses (dashed) unless independently supported.}
\label{fig:tsg}
\end{figure*}

\begin{lstlisting}[caption={Minimal temporal-scene-graph record (proposed format).},label={lst:tsg},captionpos=b]
{"pair_id": "event_x_pair_0001", "sensor": "optical",
 "timepoints": ["pre_event", "post_event"],
 "nodes": [{"id": "building_17", "type": "building",
   "bbox_pre": [x1,y1,x2,y2], "bbox_post": [x1,y1,x2,y2],
   "state_pre": "intact", "state_post": "damaged",
   "state_confidence": 0.94,
   "evidence_source": "dataset_annotation"}],
 "edges": [{"source": "building_17",
   "relation": "state_transition", "target": "damaged",
   "from_time": "pre_event", "to_time": "post_event",
   "status": "observed_or_annotated", "confidence": 0.94,
   "provenance": "damage_mask_and_label"}],
 "causal_hypotheses": [],
 "notes": "Do not infer causality from temporal order alone."}
\end{lstlisting}

\begin{table}[!htbp]
\centering
\footnotesize
\caption{\textbf{Planned graph-grounding ablation.} Same held-out events, prompts, and scoring for all rows; report graph coverage and construction errors alongside accuracy. In V3 the evaluated VLM builds the graph itself in a first prompt stage; graphs from a separate detector are kept out of the main comparison because they add an external model.}
\label{tab:ablation}
\setlength{\tabcolsep}{3pt}
\begin{tabular}{@{}llp{0.38\linewidth}@{}}
\toprule
ID & Evidence & Purpose \\
\midrule
V0 & post image & single-image baseline \\
V1 & pre + post images & bi-temporal baseline \\
V2 & oracle graph + question & diagnostic upper bound only \\
V3 & self-built graph (same VLM) & proposed method \\
V4 & corrupted graph & do graph semantics matter? (cf.\ \cond{D}) \\
V5 & reversed graph & temporal direction (cf.\ \cond{C}) \\
\bottomrule
\end{tabular}
\end{table}

\cref{fig:tsg} sketches the pipeline, \cref{lst:tsg} gives the record format, and \cref{tab:ablation} the ablation plan. The figure shows the annotation-derived (oracle) route; in the proposed self-built route (V3), steps 2--3 are carried out by the evaluated VLM from the images alone. Oracle graphs (V2) only test whether the reasoning stage can use temporal facts and must never be reported as an end-to-end result.

% ---------------------------------------------------------------------------
\FloatBarrier
\section{Run Record}
\label{app:repro}

\begin{table}[!htbp]
\centering
\footnotesize
\caption{\textbf{Run record} for the reported diagnostic (run \texttt{FULLRUN\_2026-10-12}). Fields marked ``not recorded'' were absent from the run report and will accompany the released code.}
\label{tab:runrecord}
\setlength{\tabcolsep}{3pt}
\begin{tabular}{@{}p{0.36\linewidth}p{0.6\linewidth}@{}}
\toprule
Model & OpenGVLab/InternVL3-2B-hf \\
Model revision / commit & not recorded \\
Precision / decoding & bfloat16 / greedy, $\le$10 new tokens \\
Image preprocessing & resize to 448\,px; interpolation not recorded \\
GPU & laptop GPU, 6\,GB (RTX 4050 class) \\
PyTorch / Transformers & 2.11.0 / 5.12.1 \\
CUDA / driver & not recorded \\
Seed & not recorded (greedy decoding) \\
Model calls / errors / unparseable & 72,846 / 0 / 0 \\
Wall-clock / peak memory & not recorded \\
Bootstrap & 2,000 resamples, pairs within events; interval type not recorded \\
Code commit, prompt and manifest hashes & not recorded \\
\bottomrule
\end{tabular}
\end{table}

\cref{tab:runrecord} lists what was and was not recorded for the reported run. We will release the split manifest (pair, event, sensor, disaster type, split), question manifest with exact templates, per-condition raw and parsed predictions, bootstrap outputs, and reproduction commands, subject to the DisasterM3 and model licences. We make no efficiency or deployment claim.
\FloatBarrier